\documentclass[letterpaper, 10 pt, conference]{ieeeconf}  

\IEEEoverridecommandlockouts                              

\title{\LARGE \bf
Learning Topometric Semantic Maps from Occupancy Grids 
}

\author{Markus Hiller$^{*}$, Chen Qiu$^{*}$, Florian Particke, Christian Hofmann and J{\"o}rn Thielecke
\thanks{The authors are with the Institute of Information Technology, Department of Electrical, Electronic and Communication Engineering, 
	Friedrich-Alexander-Universit{\"a}t Erlangen-N{\"u}rnberg~(FAU), Germany;
	{\mbox{Correspondence to}: \tt\small markus.hiller@fau.de}}%
\thanks{$^{*}$Both authors contributed equally to this work.}%
\thanks{This work was partly supported by the Bavarian Research Foundation BFS,
	project FORobotics (AZ-1225-16).}
}

\usepackage{amssymb}
\usepackage{graphicx}
\usepackage{subfigure}
\usepackage{amsmath}
\usepackage{nicefrac}
\usepackage{todo}
\usepackage{multirow}
\usepackage[table]{xcolor}
\usepackage{hyperref}
\usepackage{etoolbox}
\appto\UrlBreaks{\do\-\do\/}

\makeatletter
\patchcmd{\@makecaption}
{\scshape}
{}
{}
{}
\makeatother

\usepackage[absolute]{textpos}
\usepackage{xcolor}

\TPMargin{5pt}

\newcommand{\copyrightstatement}{
	\begin{textblock}{0.84}(0.08,0.015)    
		\noindent
		\footnotesize
			\centering{\color{gray}{arXiv preprint -- Accepted for presentation at the 2019 IEEE/RSJ International Conference on Intelligent Robots and Systems (IROS)}\\ \vspace{0.8mm}
			 \copyright 2019 IEEE. Personal use of this material is permitted. Permission from IEEE must be obtained for all other uses, in any current or future media, including reprinting/republishing this material for advertising or promotional purposes, creating new collective works, for resale or redistribution to servers or lists, or reuse of any copyrighted component of this work in other works.}
	\end{textblock}
}

\begin{document}
\copyrightstatement

\maketitle
\thispagestyle{empty}
\pagestyle{empty}

\begin{abstract}
Today's mobile robots are expected to operate in complex environments they share with humans. To allow intuitive human-robot collaboration, robots require a human-like understanding of their surroundings in terms of semantically classified instances. 
In this paper, we propose a new approach for deriving such instance-based semantic maps purely from occupancy grids. We employ a combination of deep learning techniques to detect, segment and extract door hypotheses from a random-sized map. The extraction is followed by a post-processing chain to further increase the accuracy of our approach, as well as place categorization for the three classes room, door and corridor. All detected and classified entities are described as instances specified in a common coordinate system, while a topological map is derived to capture their spatial links.
To train our two neural networks used for detection and map segmentation, we contribute a simulator that automatically creates and annotates the required training data. We further provide insight into which features are learned to detect doorways, and how the simulated training data can be augmented to train networks for the direct application on real-world grid maps.
We evaluate our approach on several publicly available real-world data sets. Even though the used networks are solely trained on simulated data, our approach demonstrates high robustness and effectiveness in various real-world indoor environments.
\end{abstract}

\section{INTRODUCTION}
Recent developments in broadening the field of operation for robots from secured and controllable industrial spaces to everyday environments shared with human coworkers entail the need for autonomous platforms to reliably operate in complex and populated surroundings. 
To provide assistance to humans, however, robots are in the future more than ever expected to go beyond the mode of coexistence; they should support cooperation and interaction with their environment in an intelligent and human-like manner.
This requires representations of the robot's surroundings that are able to express not only spatial properties, but actual information about the objects' identities~\cite{Cadena_SLAMSummary2016}, and some kind of abstraction level that is also understandable to humans~\cite{Bonanni_humanRobotInteraction2013}. \par 
\begin{figure}[t!]
	\centering
	\includegraphics[scale=0.35, trim={0cm 0cm 0cm -0.5cm}, clip]{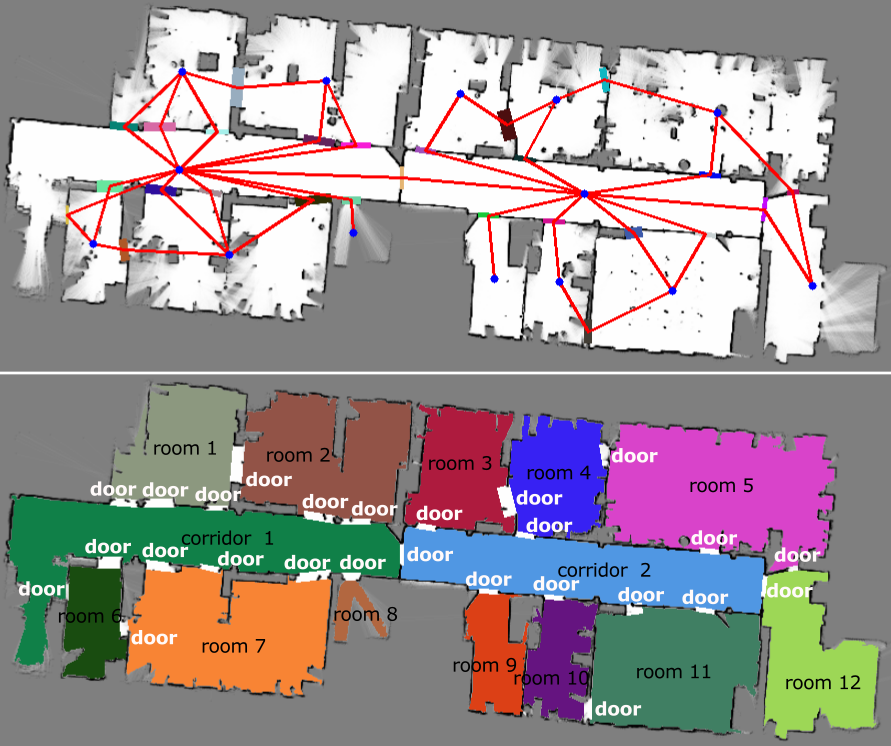}
	\vskip -0.15cm
	\caption{Result of our approach evaluated on the real-world data set of building FR079 at the University of Freiburg. (top)~Detected doors and topological links retrieved by using our proposed cascaded combination of CNN and segmentation network with additional post processing methods. (bottom)~Inferred place labels and spatial dimensions after semantic instance segmentation and place categorization.}
	\label{fig:results_motivation}
	\vskip -0.6cm
\end{figure}
While Simultaneous Localization and Mapping (SLAM) algorithms have demonstrated a recent development towards incorporation of context information (e.g.~\cite{Bowman_SemanticSLAM-DA2017}), navigation and path planning still mainly rely on occupancy grid maps first introduced by Moravec and Elfes~\cite{MoravecAndElfes_OccGridMap1985,Moravec_MappingCertaintyGrid1989}.
This results in the availability of such maps for almost any environment where robots are deployed.\par
However, in terms of deriving enhanced semantic representations, this source of information that is native to most navigation tasks has mostly been neglected in previous work. 
\par
We propose that recent developments in the area of machine and deep learning are particularly suited to extract information from such representations, apart from the sole occupancy of certain areas. Our novel approach utilizes a combination of a Convolutional Neural Network (CNN) and segmentation network to generate door hypotheses on a grid cell level. Validation of the hypotheses is then performed by 
using computer vision algorithms. Based on the estimated locations of all doorways in the environment, further semantic classification into rooms and corridors is performed, and topometric maps are derived (cf. Fig.$\,$\ref{fig:results_motivation}~and~\ref{fig:architecture_approach}). 
Our contributions to the state of the art in this work include the following: 
\begin{itemize}
	\item We propose and evaluate a learning-based architecture 
	to perform semantic segmentation, place categorization and topometric mapping on occupancy grid maps.
	\item We provide insight into which spatial structures are utilized by the CNN to correctly detect doorways.
	\item We evaluate and discuss how simulated data can be augmented to train networks that can robustly be applied to real-world map data.
	\item We further contribute a simulator for creating diverse and automatically labeled training data for supervised learning algorithms working on occupancy grid maps.\footnote{Publicly available under \tt\footnotesize{\href{https://github.com/LIKERobo/SemanticMapGeneration}{https://github.com/LIKERobo/ SemanticMapGeneration}.}}
\end{itemize}
\par
The remainder of this paper is organized as follows: In Section~\ref{sec:related_work}, we briefly discuss previous work related to our approach, before introducing it in detail throughout Sections~\ref{sec:approach_assumptions}~to~\ref{sec:hypoVal_semMapping}. Our simulator for generating required training data is briefly introduced in Section~\ref{sec:simulator_training_data}. Evaluation results of our approach on real-world data sets are presented and discussed in Section~\ref{sec:eval}.
\section{Related Work}
\label{sec:related_work}
The challenge of enriching maps with semantic information has received notable attention by several authors in the past, and a number of different appraoches for modeling such data in the form of semantic maps have been proposed~\cite{Galindo_multiHierarchySemMaps2005,Zender_conceptualSpatialMap2008,Hiller_semanticWorldModel2018}. 
To identify specific structures in the environment, a variety of different techniques including learning algorithms have been applied~\cite{Bormann_RoomSegSurvey}. Buschka et al.~\cite{Buschka_VirtSens2002} describe a method to extract room-like shapes from range-scan data.
An approach for segmenting range scans into doors and walls has been introduced by Anguelov et al.~\cite{Anguelov_DoorWallObjects2004}, whereas Limketkai et al.~\cite{Limketkai_RelationalMarkovNetworks2005} apply relative Markov networks for a similar purpose. Mozos et al. propose a system for place classification based on range data using AdaBoost in~\cite{Mozos_supervisedAdaBoost2005}, and further extend their work to incorporate vision data in~\cite{Martinez_semanticMapping2006}.
Goeddel and Olson~\cite{Goeddel_learningLabelsGridCNN2016} create occupancy grids for each scan and apply a CNN to classify different places, while Suenderhauf et al.~\cite{Suenderhauf_SemMap2015} take place categorization one step further and propose a method capable of learning new semantic classes online. 
In terms of modeling spatial relations between places, Beeson et al.~\cite{Beeson_EVG2005} apply extended Voronoi graphs to infer information for generating topological representations, while Nieto-Granda et al.~\cite{NietoGranda_SemanticMapPartitioningVoronoi2010} present an approach for partioning maps in cooperation with human guides. \par
However, all of these approaches are only applicable to live sensor data and mostly require multiple information sources like laser scanners and cameras. One of the key differences of this paper is that our learning methods can directly be applied to already existing occupancy grids that are commonly used for robot navigation, thus exploiting a new source of information apart from live sensor data.
The work probably closest to ours in terms of directly working on occupancy maps is from Liu et al.~\cite{Liu_2DSemanticOccGrid2012}, where the authors present a Markov chain Monte-Carlo based method to create a semantic map abstraction and topological representation.
In contrast to this and the previous approaches, our work is inspired by the promising results that have been achieved on grayscale images through the application of CNNs~\cite{Lecun_gradientLeNet1998}, 
as well as by recent developments in the area of semantic segmentation~\cite{Badrinarayanan_segnet2015,Ronneberger_uNet2015,Lin_refinenet2017}. The concept described in this paper applies a combination of two supervised deep learning techniques and a post-processing chain directly on occupancy grid maps to perform place categorization and to derive semantic topometric maps. Opposed to most other learning-based approaches, the networks used in this work are solely trained on data created and augmented by a simulator, while still achieving robust results on real world data.

\section{Discussion of the Approach, Challenges and Underlying Assumptions}
\label{sec:approach_assumptions}
Our approach is based on the central assumption that doorways are the major entity to indicate room access and thus define the general structure of indoor environments. Since the variety of standard door widths is bound to an interval from 0.7\,m to 1.6\,m, we hypothesize that doorways are distinctive structures that can be reliably detected in occupancy grid maps.\par
\begin{figure*}[t!] 
	\centering
	\includegraphics[scale=0.46, trim={0.08cm 0.09cm 0.05cm 0.05cm}, clip]{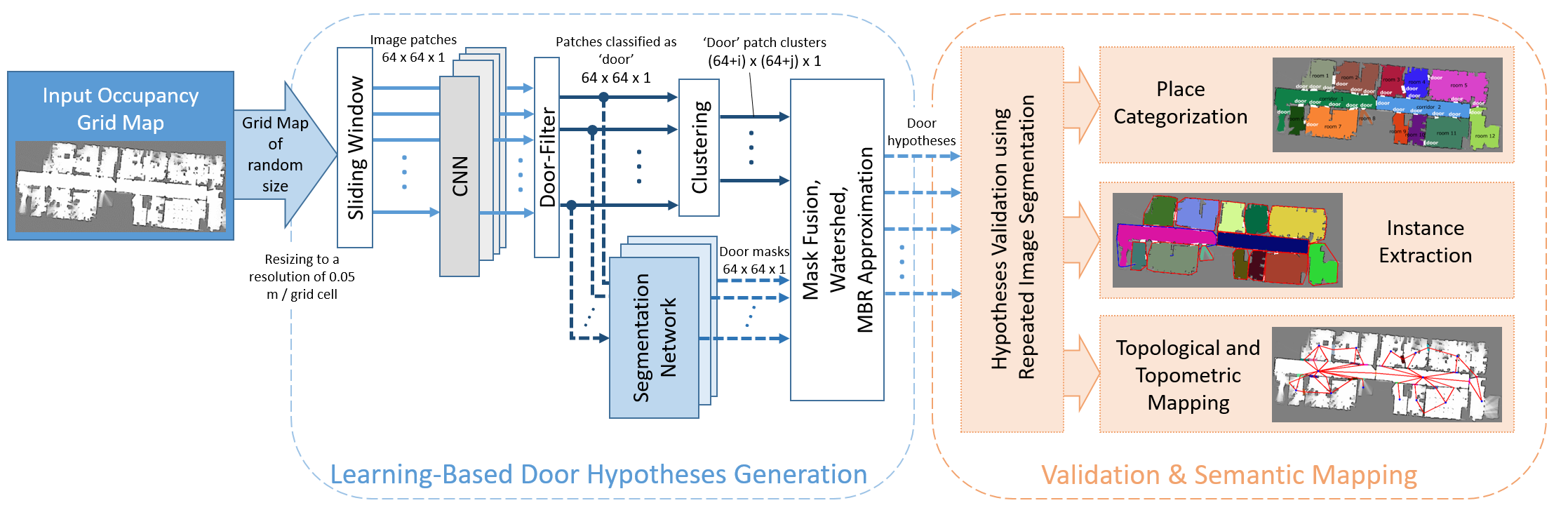}
	\vskip -0.15cm
	\caption{Architectural concept of our proposed approach.}
	\label{fig:architecture_approach}
	\vskip -0.5cm
\end{figure*}

The general architecture of our approach is depicted in Fig.$\,$\ref{fig:architecture_approach}. The concept is divided into two main parts: the generation of doorway hypotheses using deep learning techniques described in Section~\ref{sec:learning_based_hypo_gen} (depicted in blue), and the validation and semantic mapping utilizing different computer vision methods elaborated on in Section~\ref{sec:hypoVal_semMapping} (in orange).\par
Regarding the hypotheses generation, we exploit the fact that by considering the map as a whole, the underlying assumption of independent grid cells can be overcome and the displayed spatial relations can be used to derive semantic meaning based on the enclosed information. Interpreting the occupancy probabilities as a grayscale image in which each grid cell corresponds to one pixel allows the use of popular machine learning methods. The direct application of such techniques on occupancy grid maps however poses some challenges, which are discussed in the following.
\subsection{Data: Diverse and Highly Imbalanced}
Indoor environments show a high diversity regarding building sizes, shapes and dimensions of rooms as well as strongly varying structures inside the rooms due to furniture. Capturing all different variations in the form of maps would require an enormous amount of data to train neural networks. Such a data set is not publicly available at present and creation would require great effort. 
Another important challenge is that the regions of doorways contained in a map are significantly smaller than the non-doorway regions, i.e. the rest of the environment. This leads to a highly imbalanced data set which makes it very difficult to successfully train learning approaches in a straight-forward way where false positives and false negatives are treated equally. Possible ways of trying to deal with such a significant class imbalance consist of resampling the data (over-/undersampling) or specifically creating adapted loss functions.\par
We instead propose to tackle both problems at once by applying a patch-based structure using a sliding window approach. We divide the complete maps into patches of fixed size big enough to contain doorways and parts of the surrounding walls, and make the assumption that features required to recognize a doorway are sufficiently included in these patches. 
By this means, we become independent from the overall room dimensions and shapes, which significantly reduces the data complexity and allows us to create a balanced and sufficiently diverse data set through simulation for training the networks (cf. Section~\ref{sec:simulator_training_data}).
\subsection{Choice of Classifiers} 
Since we want to detect doorways in the extracted occupancy grid patches, we reformulate this problem as a binary classification based on gray scale images. However, common simple classifiers for solving such problems, like k-Nearest Neighbor~(k-NN) or Support Vector Machines~(SVM), often rely on either pixel intensities or pre-extracted features. Pixel intensities of the patches vary significantly depending on the location and surroundings of the doorway (e.g. close to a corner, objects next to the doorway, etc.) and are thus not applicable for robust classification. Distinctive features that characterize doorways are not a well studied problem. Doorways are not obviously distinctive due to one specific self-property, but rather the absence of occupied space combined with the specific surroundings of the walls and the door itself that can either be closed or open at various angles. To capture these kinds of information, we thus choose to train a CNN that is capable of extracting and learning the corresponding features to reliably classify the image patches.
To retrieve the actual doorway regions with pixel-level accuracy, we then use a segmentation network which we train to extract the cells representing the actual doorway. By only passing the image patches classified to contain doorway regions to the segmentation network as shown in Fig.$\,$\ref{fig:architecture_approach}, we avoid possible false segmentation results in areas of no interest, and the segmentation network can specialize on predicting the correct doorway masks at the pixel level.

\subsection{Choice of Overall Architecture}
Based on the assumption that doorways significantly define the structure of indoor environments, they can be used to perform semantic segmentation of the map, which is depicted in the right half of Fig.~\ref{fig:architecture_approach}.
Missed detections are at this point critical, since undetected doorways will result in a wrong overall segmentation. We therefore optimize the hypotheses generation part to minimize these false negatives by favoring a high recall rate for the CNN. To compensate possible false positives, we design a joint approach for both segmenting the map and at the same time validating the hypotheses that have been generated by the learning-based part.
Further details on how these components are realized are explained in the following two sections.
 
\section{Generation of Doorway Hypotheses}
\label{sec:learning_based_hypo_gen}
For retrieving the doorway hypotheses, a sequence of three steps is performed: region detection, segmentation and hypotheses generation, all of which will be further described in the following.
\subsection{Detection of Doorway Regions -- Bounding-Box Level}
\label{subsec:door_region_detection}
For the generation of door hypotheses, the first step is the detection of regions with high probability of containing doorways, which is realized through the following steps.
\subsubsection*{Sliding Window}
Since occupancy grid maps directly capture a discretized version of the real environment at a certain resolution, its size is linked to the environment's real-world dimensions. Conventional resizing of the input data to a fixed overall size causes great loss of information for environments of greater dimensions. We therefore instead rescale the input image to a fixed resolution of $0.05\,$m$/$grid cell, and apply a sliding window approach to split the input data into subsets of equal size. A sliding window size of $64\!\times\!64\!\times\!1$ pixels with a stride of $8$ is used at our defined input resolution to fit all standard sizes of single and double doors including frame and partial neighboring wall elements. Each extracted map part is then passed on to the CNN.
\subsubsection*{CNN-based Detection \& Filtering} 
Each partial map, or \emph{image patch}, is individually passed to the CNN which determines the class confidence for the two classes \emph{door} and \emph{background}.
We find that it is possible to achieve high performance with a relatively small network consisting of only 4 convolutional layers, connected via Rectified Linear Units~(ReLUs) and max pooling, and one fully connected layer with a sigmoid function to perform the classification. Since this network is the first and therefore crucial part of our hypotheses generation procedure, we set special emphasis on achieving a high recall rate besides a good accuracy. Possible false detections can be corrected later during a hypotheses validation step, whereas missed detections might lead to undetected doorways.
All classified patches are then passed to the filter module that rejects all non-door patches, while passing the samples with high door-confidence as well as their position in the initial map to the clustering method described in the following.
\subsubsection*{Clustering of Classified Patches}
\subfiglabelskip=0pt 
\begin{figure}[t]
	\centering
	\subfigure[][]{%
		\label{subfig:clustering_initialBoxes}%
		\includegraphics[height=3.1cm,trim={1.6cm 0.5cm 3.8cm 1.0cm},clip]{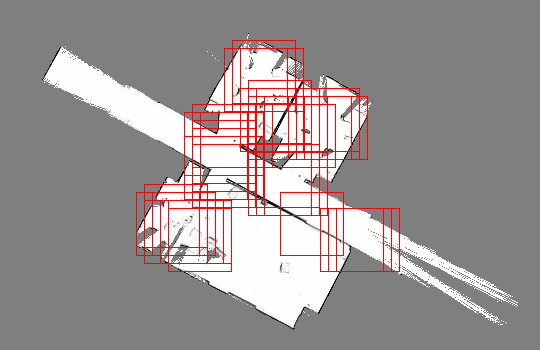}}
	\hspace{2pt}%
	\subfigure[][]{%
		\label{subfig:clustering_resultsAll}%
		\includegraphics[height=3.10cm,trim={1.2cm 0.4cm 3.0cm 0.8cm},clip]{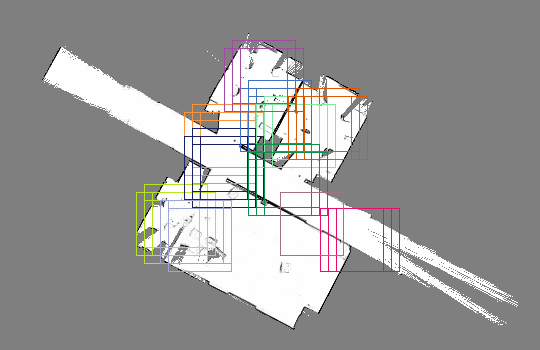}}
	\caption{CNN-based detection of door regions. \subref{subfig:clustering_initialBoxes}~Red boxes visualize the image patches retrieved from the CNN classified as regions with high probability of containing doorways. \subref{subfig:clustering_resultsAll}~Aggregation of detections to fewer hypotheses using spatial clustering.
	}
	\label{fig:clustering_patches}
	\vskip -0.6cm
\end{figure}
At this point after the filtering step, the result is a strongly reduced set of regions that have a high possibility of containing a doorway. Since the CNN is optimized for a high recall rate and a stride of 8 is used for the sliding window to extract patches of $64\!\times\!64\!\times\!1$ pixels, multiple regions with overlap are created as depicted in Fig.$\,$\ref{fig:clustering_patches}$\,$\subref{subfig:clustering_initialBoxes}. To further reduce this set to the relevant main regions, we perform clustering based on the spatial proximity. The set of patches is first sorted in descending order of their associated classification confidences. The proximity is evaluated by calculating the Intersection over Union (IoU) between the first patch in the list and all other patches. If the greatest IoU surpasses a threshold of $0.7$, the corresponding boxes are associated to the same cluster. This is repeated until all initial samples are associated to some cluster. The results are depicted in Fig.$\,$\ref{fig:clustering_patches}$\,$\subref{subfig:clustering_resultsAll}, with each cluster visualized by a different color.
\subsection{Segmentation of Detected Doorway Regions -- Pixel Level}
\label{subsec:door_region_segmentation}
Up to this point, the regions with high probability of containing doorways have been identified in a bounding-box manner (cf. Fig.$\,$\ref{fig:clustering_patches}). To enable the detection of doorways on the pixel level, we make use of a second neural network. The semantic segmentation is performed by a slightly modified version of U-Net~\cite{Ronneberger_uNet2015}, a special kind of encoder-decoder architecture originally proposed for the medical sector. The structure of the original network is adapted to fit the input data size of $64\!\times\!64\!\times\!1$ pixels and comprises a total of 19 convolutional layers and 3 max pooling operations to compress the information to feature maps of size~$8\!\times\!8\!\times\!512$ before the upscaling process, yielding good performance at still acceptable complexity. We train the network to predict a binary mask for each input patch, indicating the pixels corresponding to the doorway as visualized in Fig.$\,$\ref{fig:sim_td_concept}. By only passing the image patches classified to contain door regions to the segmentation network as shown in Fig.$\,$\ref{fig:architecture_approach}, we avoid possible false segmentation results in areas of no interest.
\subsection{Doorway Hypotheses Generation}
\label{subsec:door_hypotheses_generation}
\subfiglabelskip=0pt 
\begin{figure}[t!]
	\centering
	\subfigure[][]{%
		\label{subfig:fused_masks_orig}%
		\includegraphics[height=1.85cm,trim={0.0cm 0.0cm 0.0cm 0.0cm},clip]{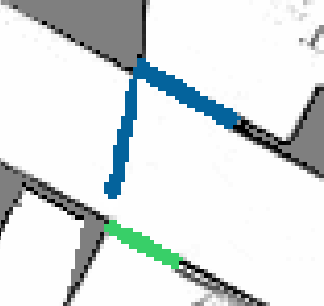}}
	\hspace{1pt}%
	\subfigure[][]{%
		\label{subfig:fused_masks_mbr_error}%
		\includegraphics[height=1.85cm,trim={0.0cm 0.0cm 0.0cm 0.0cm},clip]{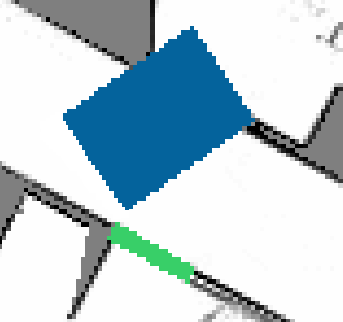}}
	\hspace{1pt}%
	\subfigure[][]{%
		\label{subfig:fused_masks_watershed}%
		\includegraphics[height=1.85cm,trim={0.0cm 0.0cm 0.0cm 0.0cm},clip] {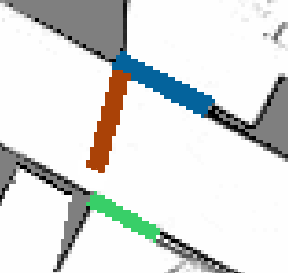}}
	\hspace{1pt}%
	\subfigure[][]{%
		\label{subfig:fused_masks_mbr_correct}%
		\includegraphics[height=1.85cm,trim={0.0cm 0.0cm 0.0cm 0.0cm},clip] {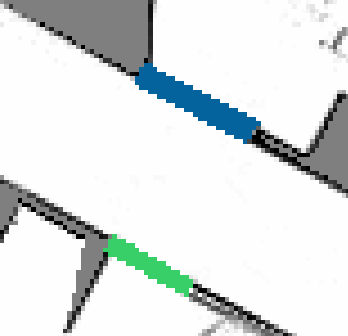}}
	\caption{Fusion and approximation of door hypotheses using minimum bounding rectangles (MBR). \subref{subfig:fused_masks_orig}~Initial fusion result of the door masks determined by the segmentation network. \subref{subfig:fused_masks_mbr_error}~Erroneous initial MBR approximation of door hypotheses. \subref{subfig:fused_masks_watershed}~Final MBR approximation of masks correctly separated by applying a combination of erosion and marker-based water shedding. 
		\subref{subfig:fused_masks_mbr_correct}~MBR of final hypotheses after validation.
	}
	\label{fig:fused_mask_postprocessing}
	\vskip -0.5cm
\end{figure}
Each image patch classified to contain a doorway has at this point one segmentation mask associated that defines the doorway on the pixel level, and is associated to one specific cluster. To retrieve the segmentation masks of the main hypotheses as depicted in Fig.$\,$\ref{fig:fused_mask_postprocessing}$\,$\subref{subfig:fused_masks_orig}, we combine all masks of the patches associated to one cluster using a pixelwise logical OR operation.
Since doorways are characterized by a rectangular shape in the real world, we further approximate the retrieved masks by using a minimum bounding rectangle~(MBR) approach as depicted in Fig.~\ref{fig:fused_mask_postprocessing}. However, if initial regions are spatially very close, multiple doorways might falsely be described by one cluster. We therefore only allow a maximum increase of $33\%$ of the area caused by the MBR approximation, since this is only intended to correct the shape. Exceeding this limit indicates that two or more masks have been falsely merged and are thus wrongly approximated by one MBR (cf. Fig.$\,$\ref{fig:fused_mask_postprocessing}$\,$\subref{subfig:fused_masks_orig} and~\subref{subfig:fused_masks_mbr_error}). 
For correction measures and under the assumption that the intersecting area of the falsely merged masks is comparably small, we interpret the fused masks as a topological surface and apply iterative erosion with a rectangular kernel of size~$2\!\times\!2$ 
to identify the local maxima. The identified regions are then used to apply marker-based water shedding to separate the falsely fused masks into more likely submasks as shown in Fig.$\,$\ref{fig:fused_mask_postprocessing}$\,$\subref{subfig:fused_masks_watershed}.
\section{Hypotheses Validation and Semantic Mapping}
\label{sec:hypoVal_semMapping}
Since the learning-based hypotheses generation approach is optimized for a high recall rate, there might exist hypotheses based on false detections. We therefore design a joint approach to assess the hypotheses based on their spatial properties while at the same time performing the instance segmentation of the map. The retrieved instances are then used as basis for place categorization and the derivation of topometric maps.
\subsection{Validation through Repeated Image Segmentation}
\label{subsec:perm_imageSeg}
We assume doorways to be the only links between different areas of the environment like rooms and corridors. We virtually close all doors by marking the corresponding areas in the grid map as occupied, and apply a graph-based segmentation algorithm introduced in~\cite{Felzenszwalb_efficient2004} to determine a total number of separated segments present in the described environment. For a successfully performed hypotheses generation process, this number directly corresponds to the total number of rooms and corridors present in the mapped scene, while the number of doorway hypotheses can be larger due to false detections. Under the assumption that only proper doorways separate the scene into distinct areas, we virtually open (unmark) one of the doorway hypotheses and repeat the segmentation process. If the hypothesis currently under validation correctly describes a doorway, the number of entities retrieved from segmentation must be by one smaller than the initially determined one. If the hypothesis is a misdetection, the number of overall segments will not change. We perform this procedure for each doorway hypothesis in a repetitive manner, and therefore refer to this approach as repeated image segmentation. An example is depicted in Fig.$\,$\ref{fig:fused_mask_postprocessing}$\,$\subref{subfig:fused_masks_watershed} and~\subref{subfig:fused_masks_mbr_correct}, where the brown hypothesis gets invalidated since it does not properly separate the environment, whereas the other two hypotheses representing actual doorways are proven valid. 
\subsection{Semantic Map Segmentation \& Place Categorization}
\label{subsec:semMapping_placeCat}
The resulting set of validated doorway hypotheses can now be used to perform the final instance segmentation to retrieve all spatial entities present in the scene. However, the semantic classes of the detected regions are at this point only known for the doorways, but not the entities retrieved via segmentation. We choose to classify those remaining regions into the two classes `corridor' and `room', following the examples of~\cite{Mozos_supervisedAdaBoost2005} and~\cite{Goeddel_learningLabelsGridCNN2016}. In contrast to these approaches, we perform the classification by developing a new measure based on the following assumptions:
\begin{enumerate}
	\item A corridor is a central linking element between rooms and thus connected to a high number of doors.
	\item The shape of a corridor is defined by showing a far less compact geometry than rooms, mostly with distinct extremities of small width.
\end{enumerate}
The number of doorways~$n_{\text{d},i}$ connected to entity~$i$ is directly deduced by performing the repeated image segmentation and associating each doorway to the two segments that become connected through its virtual opening. This is then normalized by the maximum number of doorways~$N_{\text{d},i_{max}}$ that has been assigned to any one of the entities, resulting to 
\begin{equation}
\label{eq:p_door}
p_{\text{d},i} = \frac{n_{\text{d},i}}{N_{\text{d},i_{max}}} \quad \in \left[0,\,1\right],
\end{equation} 
assuming that at least one corridor is present in the scene.\par
To quantify the shape difference between rooms and corridors, we calculate the normalized spin index~$p_{\text{s},i}$, a measure used 
in the field of landscape geography to characterize the compactness of a shape with special focus on extremities~\cite{Parent_shapeMetrics2009}. The spin index~$s_i$ is defined as the average second norm of the distances~$\mathbf{d}_i$ between all cells~$c_i$ inside entity~$i$ and their centroid, which is minimal for a circle since it denotes the most compact shape possible. To achieve independence from scaling, the spin index of an entity is normalized by the spin index~$s_{\text{EAC}}$ of a circle with the same area as this segment, resulting in a compactness measure 
\begin{equation}
\label{eq:p_spin}
p_{\text{s},i} = \frac{s_{\text{EAC}}}{s_{i}} \in \left[0,1\right], \,\,\, \text{with} \,\, s_{i}=\frac{\lVert\mathbf{d}_{i}\rVert}{n_{c_i}},\,\, s_{\text{EAC}}= 0.5\,r^2_{\text{EAC}}.
\end{equation}
A small normalized spin index~$p_{s,i}$ denotes low compactness and therefore a high probability for the entity~$i$ to be of the class `corridor'. Since both measures~\eqref{eq:p_door} and~\eqref{eq:p_spin} are defined over an interval~$[0,1]$, we retrieve our combined confidence score~$p_{\text{comb},i}$ for an entity~$i$ representing a corridor as 
\begin{equation}
p_{\text{comb},i} =  p_{\text{d},i} \cdot \left(1-p_{\text{s},i}\right)\,\, \in \left[0,\,1\right].
\end{equation}
These measures are now used to assign a place label to each of the previously extracted map segments, leading to a semantically classified representation.
Examples for the results achieved with all three measures are depicted in Fig.$\,$\ref{fig:place_categorization}, and further discussed in Section~\ref{sec:eval}.
\subfiglabelskip=0pt 
\begin{figure}[t!]
	\centering
	\subfigure[][]{%
		\label{subfig:map_topologic}%
		\includegraphics[height=3.5cm, trim={-0.2cm 0.0cm 0.0cm 0.0cm}, clip]{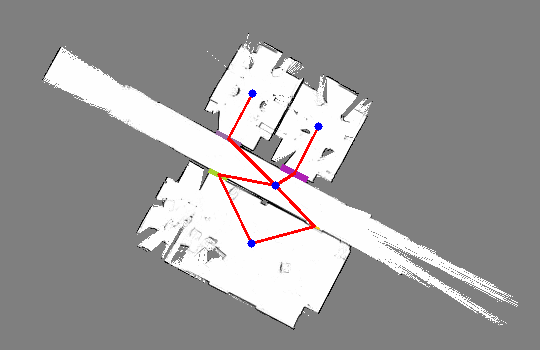}}
	\hspace{3pt}%
	\subfigure[][]{%
		\label{subfig:map_topometric}%
		\includegraphics[height=3.5cm, trim={0.0cm 0.5cm 0.0cm 0.0cm}, clip]{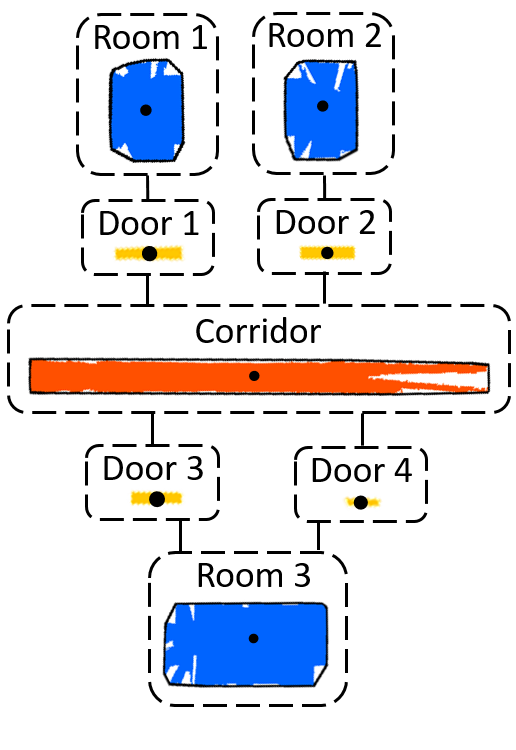}}
	\vskip -0.3cm
	\caption{Derivation of an instance-based topometric map. \subref{subfig:map_topologic}~Topologic map structure overlayed to the original occupancy grid map. \subref{subfig:map_topometric}~Instance-based topometric map with each segment described by reference point, shape approximation and topologic links.
	}
	\label{fig:map_topometric_compil}
	\vskip -0.3cm
\end{figure}
\subsection{Topometric Map}
\label{subsec:topologic_map}
For high-level navigation and efficient planning tasks, topological representations are often preferred to metric maps~\cite{Thrun_LearningTopoMetric1998}, however with the downside of neglecting the detailed metric information. 
Due to the performed instance segmentation and association of all doorways to the neighboring rooms and corridors, we provide the possibility to directly derive a map not only specifying the elements' relations in the form of a topological structure (cf. Fig.$\,$\ref{fig:map_topometric_compil}$\,$\subref{subfig:map_topologic}), but to also capture the spatial dimensions of each linked element in the form of a so called topometric map (cf. Fig.$\,$\ref{fig:map_topometric_compil}$\,$\subref{subfig:map_topometric}), similar to~\cite{Kostavelis_SemanticMapSurvey2015}. To support instance-based environment models like the one proposed in~\cite{Hiller_semanticWorldModel2018}, we additionally describe each extracted region by a reference point (e.g. the centroid) and approximate the region's shape by determining the convex hull as depicted in Fig.$\,$\ref{fig:map_topometric_compil}$\,$\subref{subfig:map_topometric}. This allows a very compact representation capturing all relevant entities. 

\begin{figure}[t!]
	\centering
	\includegraphics[scale=0.37, trim={0cm 0.4cm 0.8cm 0.0cm}, clip]{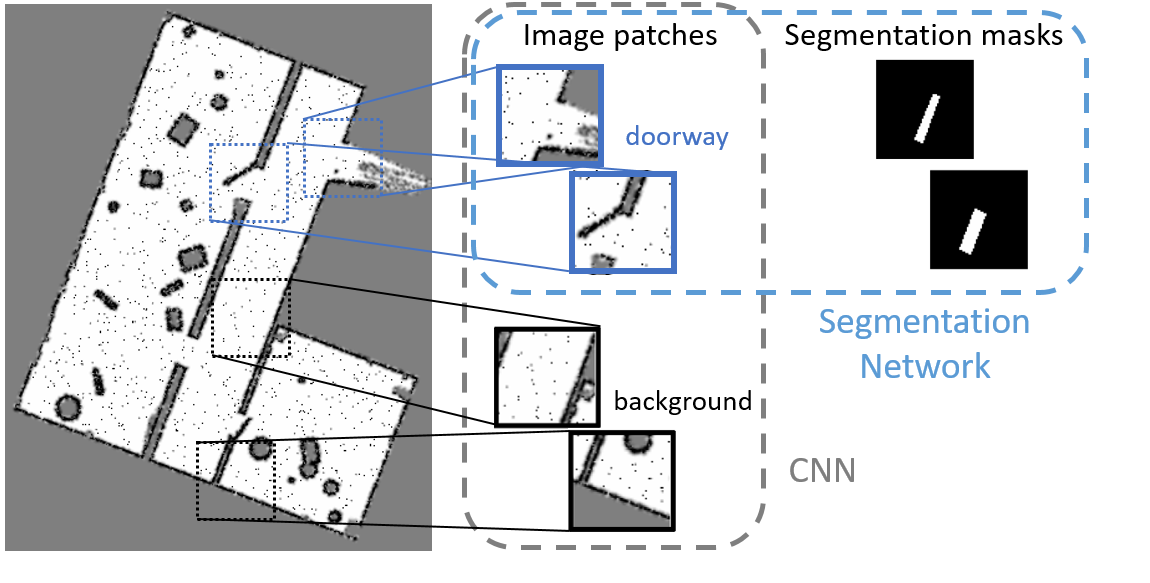}
	\vskip -0.15cm
	\caption{Generation of training data with our simulator. An occupancy grid map is generated, image patches of certain size are automatically extracted and labeled for training the CNN, and segmentation masks are additionally created for all doorway patches to train the segmentation network.}
	\label{fig:sim_td_concept}
	\vskip -0.6cm
\end{figure}
\section{Generation of Training Data}
\label{sec:simulator_training_data}
\begin{figure}[t]
	\centering
	\subfigure[][]{%
		\label{subfig:sim_patch_ideal}%
		\includegraphics[height=1.85cm,clip]{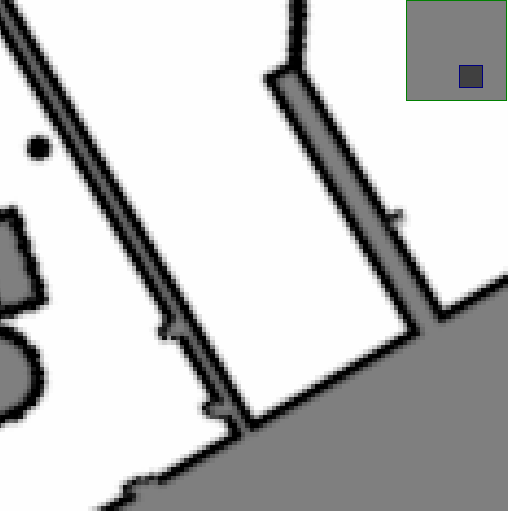}}
	\hspace{3pt}%
	\subfigure[][]{%
		\label{subfig:sim_patch_impEdges}%
		\includegraphics[height=1.85cm,clip]{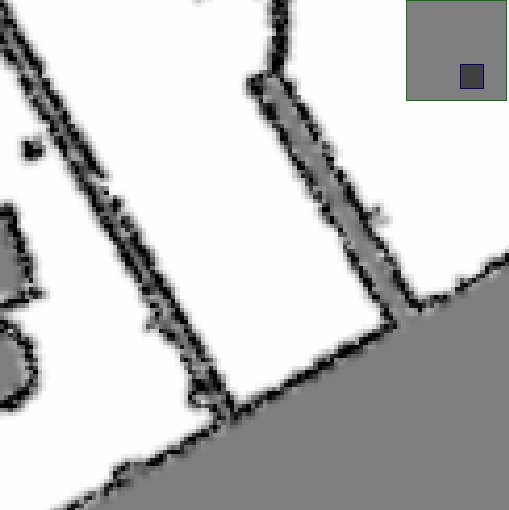}}
	\hspace{3pt}%
	\subfigure[][]{%
		\label{subfig:sim_patch_areaNoise}%
		\includegraphics[height=1.85cm,clip]{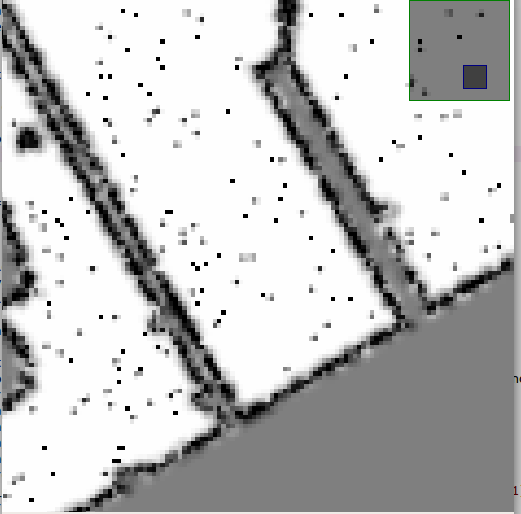}}
	\hspace{3pt}%
	\subfigure[][]{%
		\label{subfig:sim_patch_realMap}%
		\includegraphics[height=1.85cm,clip]{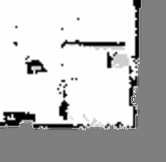}}
	\vskip -0.15cm
	\caption{Different stages of simulator-based generation of occupancy maps. \subref{subfig:sim_patch_ideal}~Initial idealized map. \subref{subfig:sim_patch_impEdges}~Modeled measurement uncertainties result in imperfections on the edges. \subref{subfig:sim_patch_areaNoise}~Incorporation of sporadic measurement errors and other artefacts. \subref{subfig:sim_patch_realMap}~Section of a map created with real-world data.
	}
	\label{fig:simulator_training_data}
	\vskip -0.6cm
\end{figure}

One of the challenges for the application of deep learning approaches to occupancy grid maps discussed in Section~\ref{sec:approach_assumptions} is the lack of sufficient publicly available data sets for training purposes. We therefore developed and contribute a grid map generator that creates a diverse set of labeled map patches and can be used for any supervised learning method working on occupancy grids.
The generation of training samples is divided into two major steps: the creation of full occupancy grid maps, and the extraction of annotated training samples as visualized in Fig.$\,$\ref{fig:sim_td_concept}.
The map creation is based on an arrangement of random-sized rooms with a variable number of doors, opened at a random angle, around one central corridor. 
To emulate a more realistic setting, shapes representing objects like furniture are added inside the unoccupied spaces.
Finally, the map is rotated at a random angle to account for the fact that the orientation of real occupancy grid maps is not fixed. 
All parameters are randomly chosen inside intervals whose bounds can either be specified or are constrained by spatial relations. 
This procedure generates maps like the one partially displayed in Fig.$\,$\ref{fig:simulator_training_data}$\,$\subref{subfig:sim_patch_ideal}. However, this represents an idealized version of a map. 
The comparison to the map based on real-world data depicted in Fig.$\,$\ref{fig:simulator_training_data}$\,$\subref{subfig:sim_patch_realMap} shows that this vanilla approach is not suited for the generation of a robust and diverse data set that models the characteristics of the real-world data.
The simulator is therefore extended to model the effects of uncertainties and measurement errors for different levels of noise. We incorporate two methods to augment the data: additive white Gaussian noise and a noise type specifically designed to resemble the visually perceivable imperfections present in real maps, dubbed \emph{combined noise}. 
For the combined noise, 
a combination of salt and pepper noise is applied along the edges to model non-Gaussian measurement errors, e.g. caused by differences in surface reflection. Gaussian noise is then added to smooth the effects and to model Gaussian measurement uncertainty inherent in the sensors used for recording. The result and influence onto the edge structure is depicted in Fig.$\,$\ref{fig:simulator_training_data}$\,$\subref{subfig:sim_patch_impEdges}.
At last, pepper noise is added to the unoccupied space to model sporadic measurement errors as well as artefacts like inconsistent measurements caused by dynamic objects present in real maps. The qualitative comparison of the basic characteristics of the resulting generated map depicted in Fig.$\,$\ref{fig:simulator_training_data}$\,$\subref{subfig:sim_patch_areaNoise} and the real map shown in$\,$\subref{subfig:sim_patch_realMap} leads to the assumption of a high resemblance and thus suitability for training purposes. This assumption is quantitatively supported by the results of our analysis presented in Section~\ref{sec:eval}.

\section{Evaluation}
\subfiglabelskip=0pt 
\subfiglabelskip=0pt 
\begin{figure}[t]
	\centering
	\subfigure[][]{%
		\label{subfig:learn_door_sim}%
		\includegraphics[height=1.85cm,clip]{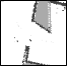}}
	\hspace{3pt}%
	\subfigure[][]{%
		\label{subfig:learn_door_sim_heat}%
		\includegraphics[height=1.85cm,clip]{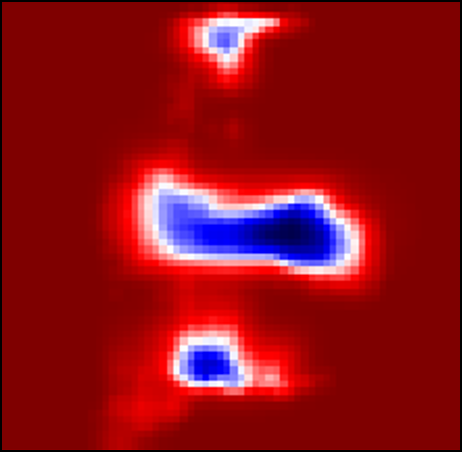}}
	\hspace{3pt}%
	\subfigure[][]{%
		\label{subfig:learn_door_real}%
		\includegraphics[height=1.85cm,clip]{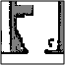}}
	\hspace{3pt}%
	\subfigure[][]{%
		\label{subfig:learn_door_real_heat}%
		\includegraphics[height=1.85cm,clip]{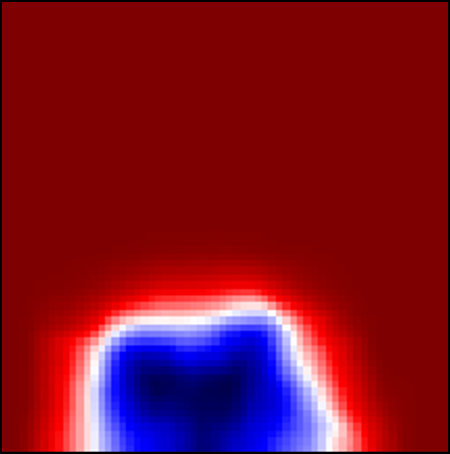}}
	\vskip -0.15cm
	\caption{Areas of relevant features learned by the CNN for the classification of doorways. \subref{subfig:learn_door_sim} and \subref{subfig:learn_door_real} depict map patches extracted from real-world data sets containing a doorway, while \subref{subfig:learn_door_sim_heat} and \subref{subfig:learn_door_real_heat} show the associated heat maps representing the importance of the corresponding areas for the classification result (blue denotes high importance).
	}
	\label{fig:learn_door_features}
	\vskip -1.2cm
\end{figure}
\label{sec:eval}
\begin{figure*}[thb]
	\centering
	\subfigure[][]{%
		\label{subfig:eval_pub_fr52}%
		\includegraphics[height=3.7cm, trim={0.0cm 0.0cm 0.0cm 0.0cm}, clip]{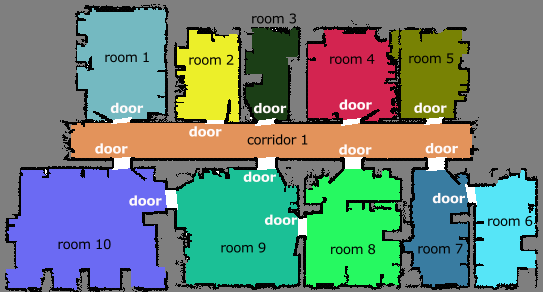}}
	\hspace{1pt}%
	\subfigure[][]{%
		\label{subfig:eval_pub_fr101}%
		\includegraphics[height=3.7cm, trim={0.0cm 0.5cm 0.0cm 0.0cm}, clip]{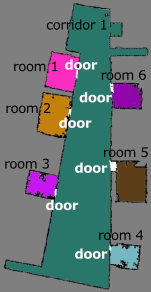}}
	\hspace{1pt}%
	\subfigure[][]{%
		\label{subfig:eval_pub_intel}%
		\includegraphics[height=3.7cm, trim={0.0cm 0.0cm 0.0cm 0.0cm}, clip]{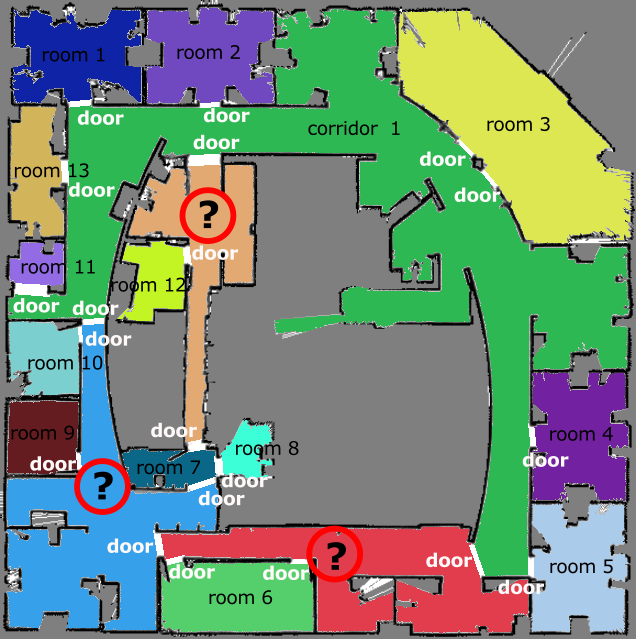}}
	\hspace{1pt}%
	\subfigure[][]{%
		\label{subfig:eval_pub_intelheat}%
		\includegraphics[height=3.7cm, trim={0.0cm 0.0cm 0.0cm 0.0cm}, clip]{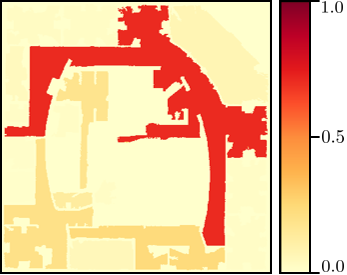}}
	\vskip -0.15cm
	\caption{Evaluation on three real-world data sets: two from the University of Freiburg and the Intel building. \subref{subfig:eval_pub_fr52}~Segmentation and classification results on building FR052 and \subref{subfig:eval_pub_fr101}~on building FR101. \subref{subfig:eval_pub_intel}~Segmentation and classification results on the Intel building data set. Uncertain classifications marked by circled question marks. \subref{subfig:eval_pub_intelheat}~Place categorization results for the class `corridor' displayed as heatmap, applied to the Intel building data set.
	}
	\label{fig:eval_pub_fr}
	\vskip -0.5cm
\end{figure*}
To provide insight into the results of the proposed approach, we first analyze what the CNN learns and how data augmentation helps to achieve robust classification results. To assess the performance of the proposed overall approach, we evaluate the techniques on several real-world data sets. \vspace{-4mm} \footnote{Provided by C. Stachniss, D. Haehnel and others, available under \tt\footnotesize{\href{http://www.ipb.uni-bonn.de/datasets/}{http://www.ipb.uni-bonn.de/datasets/}.}}
\begin{figure}[thb]
	\centering
	\subfigure[][]{%
		\label{subfig:eval_augm_CNN}%
		\includegraphics[height=5.0cm, trim={0.0cm 0.0cm 0.0cm 0.0cm}, clip]{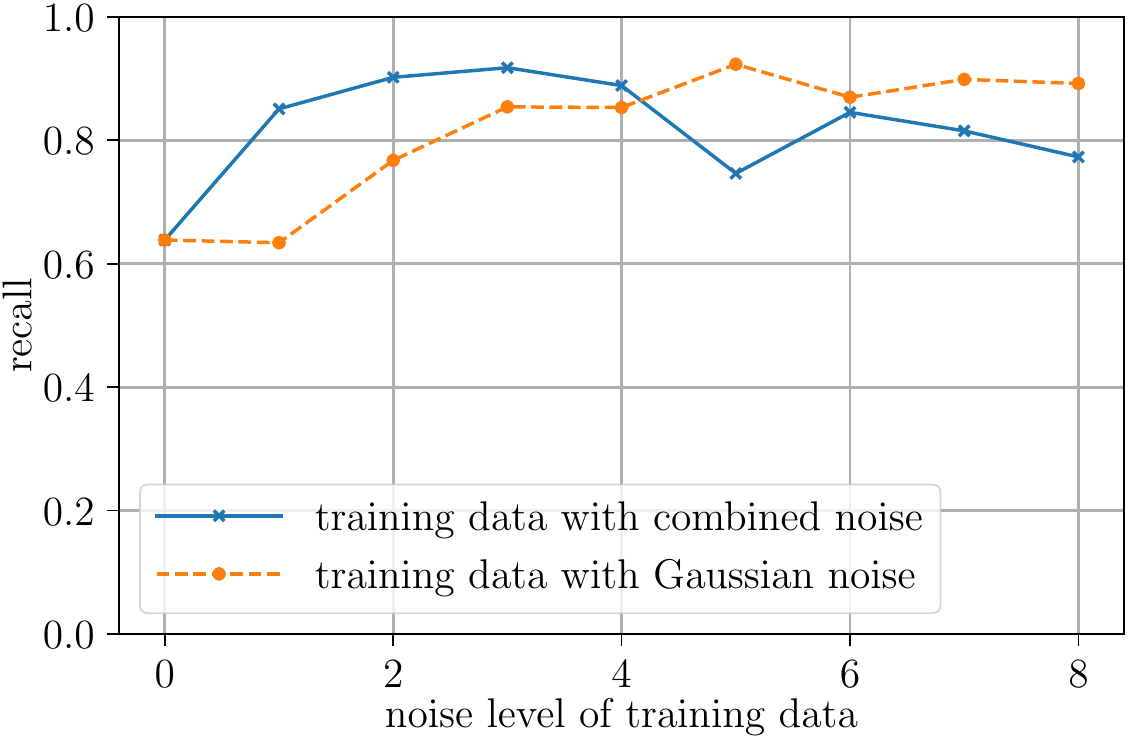}}
	\vskip -0.15cm
	\subfigure[][]{%
		\label{subfig:eval_augm_SegUNet}%
		\includegraphics[height=5.0cm, trim={0.0cm -0.35cm 0.0cm 0.0cm}, clip]{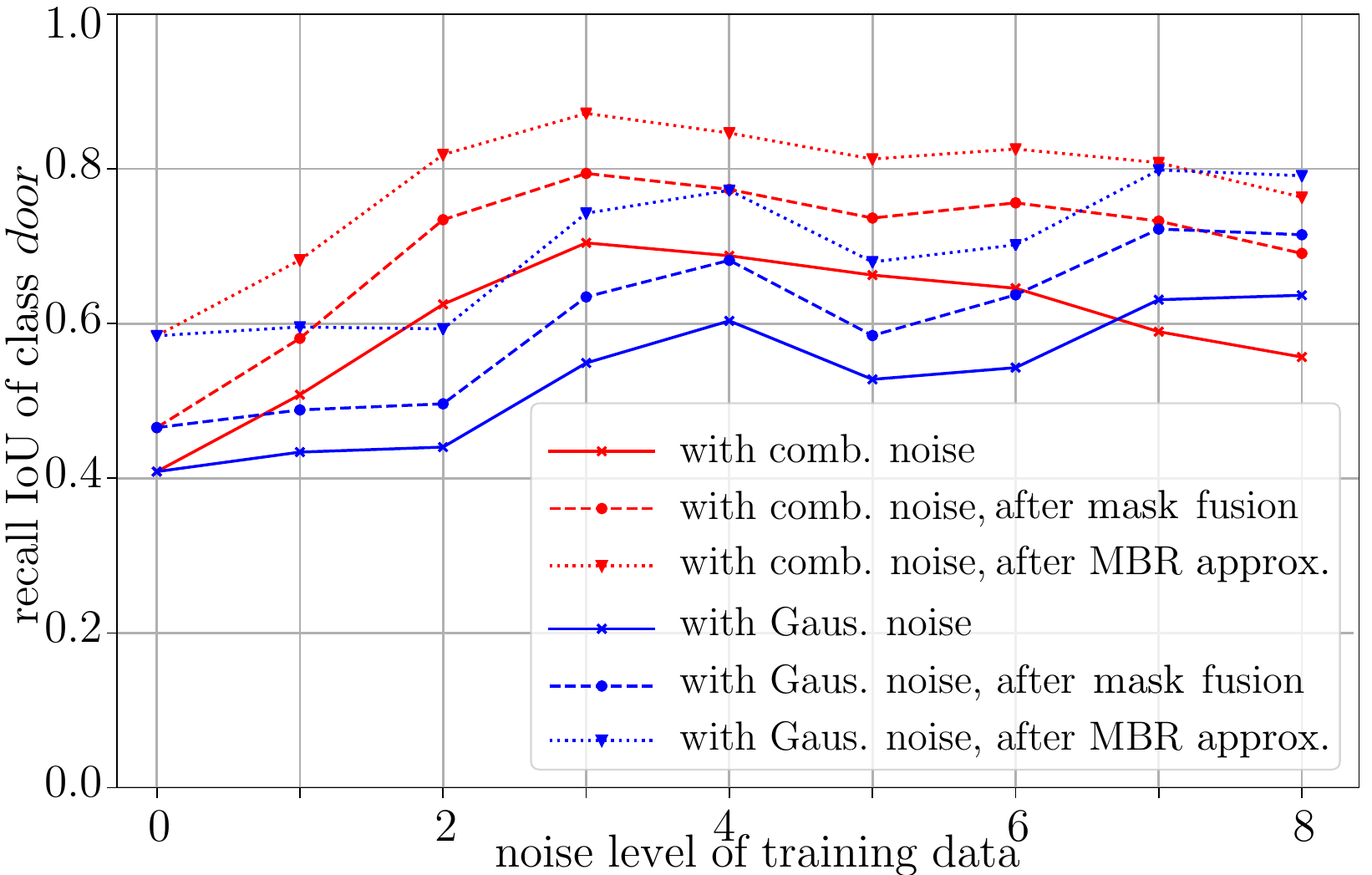}}
	\vskip -0.25cm
	\caption{Evaluation results comparing two approaches for data augmentation using different types of noise. Both networks have been trained on simulated data and are tested on real-world data. \subref{subfig:eval_augm_CNN}~Classification results achieved by the CNN. \subref{subfig:eval_augm_SegUNet}~Results achieved by the segmentation network. 
	}
	\label{fig:eval_augm}
	\vskip -0.7cm
\end{figure}
\subsection{What does the CNN learn?}
An important question is if a trained classifier can generalize enough to achieve robust results on unseen data. In our case, it is important to gain insight into which kind of structure in the map is important for the CNN to classify a region as doorway. We use the approach of~\cite{Zeiler_Visualizing2014} to visualize which regions are used by the CNN to perform the classification by partially occluding specific image regions. As depicted in Fig.~\ref{fig:learn_door_features}, the CNN is 
able to perform correct classification only if certain areas of the map (depicted in blue) are available. The CNN mainly relies on the neighboring wall parts but also on free space to extract features that define if a provided patch is classified as doorway. 
\subsection{How does simulated data have to be augmented?}
One of the key questions that arises if real-world data is not sufficiently available is whether networks can be trained on simulated data that is augmented in a way that they can still achieve robust performance on real-world data.
To answer this question in the context of our approach, we incorporate two different methods to augment the simulated data in order to model errors present in real-world maps (cf. Section~\ref{sec:simulator_training_data}). We therefore train our networks on the same set of 60000 generated samples, however each time augmented by a different level of Gaussian or combined noise with increasing intensities. We evaluate the performance for each setting based on 1090 patches extracted from 12 real-world data sets and labeled by hand. The results regarding the evaluation of the CNN depicted in Fig.~\ref{fig:eval_augm}\,\subref{subfig:eval_augm_CNN} show that the recall can be increased from approximately 0.63 without augmentation (level 0) to over 0.9 by using any of the two proposed methods for augmentation. Using the handcrafted combined noise during training achieves its best results for noise level~3, while the Gaussian variant peaks for noise level~5. This shows that for training the CNN, both are generally suited. Since the training of our segmentation network relies on the same data set (doorway samples in combination with appropriate pixelwise masks), we analyze the influence of both augmentation methods regarding the segmentation performance. Even though the mean intersection over union (mIoU) is widely used for segmentation problems, it focuses on the accuracy of the predicted masks and equally penalizes false positives and negatives. In this work however, false negatives are far more critical since they can result in an only partially detected doorway that does not fully cover the actual doorway region, leading to problems in the hypotheses validation procedure. Analogously to the recall of classification, we define the recall-IoU as a metric considering the percentage of  actual positives that are correctly detected. The results depicted in Fig.~\ref{fig:eval_augm}\,\subref{subfig:eval_augm_SegUNet} show that using the combined noise for data augmentation clearly outperforms the Gaussian approach over most noise levels, peaking at level 3 with a mean recall-IoU of approximately 0.7. Through applying the fusion of masks and approximation using MBR introduced in Section~\ref{sec:hypoVal_semMapping}, the mean recall-IoU can be increased up to 0.87.\par
Summarizing the results of both analyses, choosing the combined noise for data augmentation yields very promising results for both classification and segmentation networks and allows a significant increase in performance when applied to real-world data. For further details regarding the exact composition of the noise levels, the reader is referred to the documentation of the simulator$^1$.

\subsection{Evaluation on Public Real-World Data Sets}
\label{subsec:eval_real}
This section is focused on the qualitative evaluation of the overall approach on several publicly available data sets.
\subsubsection*{Doorway Detection and Instance Segmentation}
We first concentrate on the correct detection of doorways (labeled as `door' and displayed in white color) and the instance segmentation of the different areas (visualized by varied colors) of the underlying data sets.
Our approach achieves a very clear separation of the $11$ regions on the data set FR052 depicted in Fig.$\,$\ref{fig:eval_pub_fr}$\,$\subref{subfig:eval_pub_fr52} after detecting $12$ valid doorways.
Results of similar performance are achieved on the data sets of building FR101 shown in Fig.$\,$\ref{fig:eval_pub_fr}$\,$\subref{subfig:eval_pub_fr101} and of building FR079 visualized in Fig.$\,$\ref{fig:results_motivation}, even though there might be a non-detected doorway near the top-right corner in FR052, and some additionally detected doorways in FR079.
A challenge for our approach poses the Intel data set depicted in Fig.$\,$\ref{fig:eval_pub_fr}$\,$\subref{subfig:eval_pub_intel} with its very unique structure. Some of the very narrow passages are recognized as doorways based on the strong resemblance they have in the patches used for classification (e.g. between `room~10' and~`11'), whereas other parts show probably doors defined by relatively wide spaces between walls which were not modeled in our training set (e.g. on top part of `corridor~1' between `room~2' and~`3'). This results in some falsely segmented instances. The overall performance on this specific data set is nevertheless promising since all rooms with distinctive doorways are correctly segmented. 
\begin{figure}[t]
	\centering
	\subfigure{%
		\label{subfig:placeCat_doorIntensity}%
		\includegraphics[height=1.7cm, trim={-0.0cm 0.0cm 0.0cm 0.0cm}, clip]{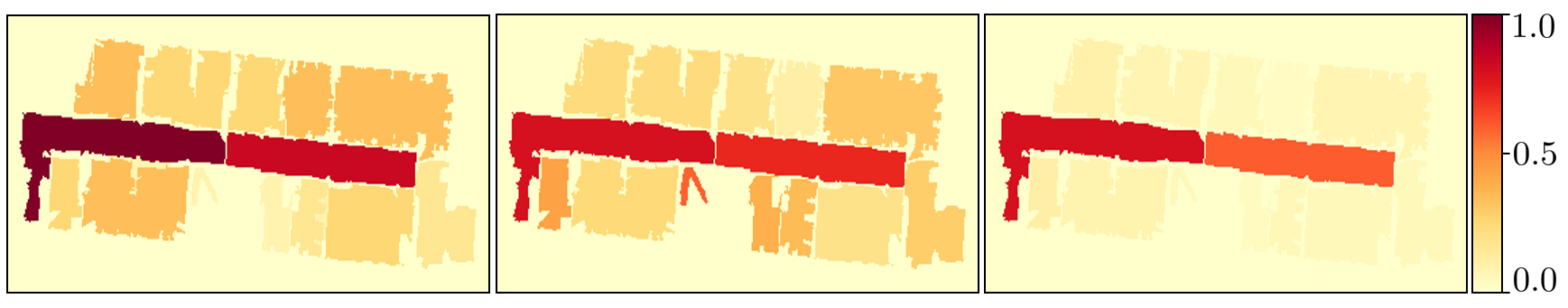}}
	\vskip -0.3cm
	\caption{Place categorization results for data set from Univ. Freiburg, building FR079. Depicted are the corridor confidences based on (left)~doorway intensity criterion~$p_{\text{d}}$, (middle)~normalized spin index criterion~$p_{\text{s}}$, and (right)~combined criterion~$p_{\text{comb}}$.
	}
	\label{fig:place_categorization}
	\vskip -0.5cm
\end{figure}
\subsubsection*{Place Categorization}
Results for both components (doors and normalized spin) of the measure as well as the resulting one used for classifying the corridor and rooms is depicted in Fig.$\,$\ref{fig:place_categorization} for the data set FR079. Both criteria individually identify the corridor, yielding a clear place categorization result. Similar results are achieved for the data sets FR052 and FR101 (not depicted).
The Intel data set again poses a challenge also for the categorization task, with the resulting heatmap for the corridor classification depicted in Fig.$\,$\ref{fig:eval_pub_fr}$\,$\subref{subfig:eval_pub_intelheat}. While one part of the actual corridor is assigned a high confidence and most of the rooms are clearly categorized, the remaining parts of the corridor are segmented in a sub-optimal way due to the aforementioned reasons, leading to inconclusive categorization results emphasized through the question marks in Fig.$\,$\ref{fig:eval_pub_fr}$\,$\subref{subfig:eval_pub_intel}.
\subsection{Comparison of segmentation results to other state of the art approaches}
Since our approach is capable of operating on partial maps with unresolved borders, it is also applicable to be operated on correspondingly mapped live data, even though this has not yet been extensively tested.
Compared to the work presented in~\cite{Mozos_supervisedAdaBoost2005} and~\cite{Goeddel_learningLabelsGridCNN2016}, we achieve classification results for the same three classes but with distinctively cleaner segmentation and fewer artefacts. Even though other approaches like~\cite{Suenderhauf_SemMap2015} make use of more sophisticated methods, they also produce artefacts in the maps due to the fact that the class of each cell is updated based on the area the laser scanner observes at the time of place categorization, resulting in sensor frequency and viewpoint dependency. Due to our underlying assumption of the door as interconnecting instance and the thereon based segmentation procedure, we are independent of these properties. However, since we are only using the information of occupancy grid maps in this approach, Suenderhauf et al.~\cite{Suenderhauf_SemMap2015} are able to label the places in a much greater variety. In future work, the approaches could be combined for achieving both an increased variety of place labels and clean segmentation results.
\section{CONCLUSION}
In this work, we introduced and evaluated a novel learning-based approach for deriving semantically annotated, instance-based topometric environment representations directly from occupancy grid maps of random sizes. We provided insight into which structures are important for a CNN to correctly classify doorways. We further presented two solutions to augment simulated data in a way that it can be used for training networks that achieve robust results on several evaluated real-world data sets. 

\bibliographystyle{IEEEtran}
\bibliography{IEEEabrv,literatur}

\end{document}